%% file: main.tex
\documentclass[9pt]{isprs} 
\usepackage{setspace}
\usepackage{geometry} 
\usepackage{epstopdf}
\usepackage{amsfonts}
\usepackage{mathtools}
\usepackage{algorithm}
\usepackage{algpseudocode}

\usepackage{tabularx}
\usepackage{booktabs} 
\usepackage{siunitx}
\robustify\bfseries  
\usepackage{multirow} 
\usepackage{array}
\usepackage{multicol}
\usepackage{dcolumn}
\usepackage{microtype}
\usepackage{url}
\usepackage{makecell} 
\usepackage{graphicx}
\usepackage[dvipsnames]{xcolor}
\usepackage{subcaption} 
\usepackage[labelsep=period]{caption}  
\usepackage[british]{babel} 
\usepackage[hang]{footmisc}
\usepackage[colorlinks=true, linkcolor=blue, breaklinks=true, urlcolor=blue]{hyperref}
\usepackage[capitalize,noabbrev,nameinlink]{cleveref} 

\newtheorem{definition}{Definition}

\usepackage{enumerate}
\usepackage[round, authoryear]{natbib}
\usepackage{tumabbrev}
\usepackage{tikz}
\usetikzlibrary{calc, positioning}
\usepackage{pgfplots,pgfplotstable}
\usepgfplotslibrary{groupplots}
\pgfplotsset{compat=newest}
\usepgfplotslibrary{fillbetween}
\usepackage[
 acronym,
 automake,
]{glossaries-extra}
\makeglossaries

\setabbreviationstyle[acronym]{long-short-user}
\renewcommand*{\glsxtruserparen}[2]{
  \glsxtrfullsep{#2}%
  \glsxtrparen
   {#1\ifglshasfield{\glsxtruserfield}{#2}{;\xspace%
     \expandafter\citealp\expandafter{\glscurrentfieldvalue}%
   }{}%
   }%
}

\glsdefpostdesc{acronym}{
 \ifglshasfield{\glsxtruserfield}{\glscurrententrylabel}%
 {~\expandafter\citep\expandafter{\glscurrentfieldvalue}}%
 {}%
}

\glsdisablehyper
\defglsentryfmt[acronym]{\ifglsused{\glslabel}{\glsgenentryfmt}{\emph{\glsgenentryfmt}}}

\glsdisablehyper
\defglsentryfmt{\ifglsused{\glslabel}{\glsgenentryfmt}{\emph{\glsgenentryfmt}}}

\input{./acronyms}

\usepackage{xspace}
\usepackage{relsize}  

\newcommand{\dataset}[1]{\textsc{#1}\xspace}
\newcommand{\class}[1]{{\smaller{\texttt{#1}}}\xspace}%
\newcommand{\EuroCrops}{\dataset{EuroCrops}}
\newcommand{\EuroCropsML}{\dataset{EuroCropsML}}
\newcommand{\meadow}{\class{grassland grass}}
\newcommand{\wheat}{\class{wheat}}
\newcommand{\parsley}{\class{parsley}}

\newcommand{\colorblue}{\color{blue}}

\begin{document}

\title{Mind the Gap: Bridging Prior Shift in Realistic Few-Shot Crop-Type Classification}
\date{November 2025}

\author{%
Joana Reuss\textsuperscript{1,}\thanks{Corresponding author} ,
Ekaterina Gikalo\textsuperscript{1},
Marco K\"orner\textsuperscript{1,2,3}
}
\address{%
  \textsuperscript{1 } Technical University of Munich (TUM), TUM School of Engineering and Design, Department of Aerospace and Geodesy,\\
  Chair of Remote Sensing Technology, 80333 Munich, Germany\\
  \textsuperscript{2 } Technical University of Munich (TUM), Munich Data Science Institute (MDSI), 85748 Garching, Germany\\
  \textsuperscript{3 } ELLIS Unit Jena, University of Jena, 07743 Jena, Germany
}
\abstract{
Real-world agricultural distributions often suffer from severe class imbalance, typically following a long-tailed distribution.
Labeled datasets for crop-type classification are inherently scarce and remain costly to obtain.
When working with such limited data, training sets are frequently constructed to be artificially balanced---in particular in the case of few-shot learning---failing to reflect real-world conditions.
This mismatch induces a shift between training and test label distributions, degrading real-world generalization.
To address this, we propose \gls{gl:dipa}, a novel method that simulates an unknown label distribution skew of the target domain proactively during model training. 
Specifically, we model the real-world distribution as \emph{Dirichlet}-distributed random variables, effectively performing a prior augmentation during few-shot learning.
Our experiments show that \gls{gl:dipa} successfully shifts the decision boundary and stabilizes the training process by acting as a dynamic feature regularizer.
}


\maketitle


\section{Introduction}\label{sec:intro}
In light of the increasing risk factors associated with food security, accurate agricultural monitoring is becoming increasingly crucial.
While \gls{gl:ml} methods have achieved state-of-the-art performance on multi-spectral (time series) crop-type data \citep{QI23:deeplcropclassification,Saini18:cropclassificationRFSVM}, their reliability in real-world scenarios remains critically hindered by data scarcity and distributional shifts.

Real-world label distributions in crop-type classification are often highly skewed.
For instance, common crops like \wheat dominate the landscape, while rare ones like \parsley are heavily underrepresented.
Compounding this issue, the high costs and labor required to acquire accurate crop-type labels often limit the available data to only a few examples per class, making \gls{gl:fsl} a practical approach for this domain.
Frequently, \gls{gl:fsl} is equated with the concept  of \glsxtrlong{gl:mtl} but it can also simply be used in order to describe the low-data constraint in isolation.
We use \gls{gl:fsl} to refer to the latter, noting that recent work has shown that regular transfer learning via pretraining and fine-tuning can receive competitive results to complex meta-learning algorithms \citep{Reuss25:MTLSSL,Chen18:fewshotclassification}.
However, labeled training (support) datasets in \gls{gl:fsl} are often constructed with a balanced label distribution.
This reflects an idealized scenario where, given data scarcity, samples are deliberately collected to stabilize learning and ensure a fair representation across all classes.
In order to reflect realistic scenarios, the standard practice of using a balanced \gls{gl:fsl} test (query) set has been criticized as being unrealistic, with studies recommending the use of arbitrary and imbalanced test sets \citep{Veilleux21:dirichletevaluation,Ochael23:classimbalanceFSL,Mohammadi24:fewshotcropmappingdirichlet}.
Consequently, the training class prior $p_\text{train}(y)$ is not representative of the real-world test prior $p_\text{test}(y)$.
As a result, during the testing or inference phase, the model is exposed to a distributional shift, leading it to learn a strong, incorrect bias.
This typically results in poor generalization performance, especially when the dataset is small or long-tailed \citep{Reuss25:MTLSSL}.

While most existing methods address such label or prior shifts \textit{post-hoc} after training by correcting the predicted class probabilities at inference time \citep[\cf \cref{sec:relatedwork}]{Lipton18:BlackBox,Kluger21:FPSA}, we propose to model the prior uncertainty proactively during the training process.
Specifically, we leverage the 
\gls{gl:DirDist}
to sample a vast range of class distributions.
This exposes the model to various realistic label distributions, ultimately leading to a classifier with superior robustness to prior shift during inference without any knowledge of the actual test skew.
This is particularly critical when generalizing from a few labeled samples.

The main contributions of this work are:
\begin{enumerate}
    \item \textbf{Prior-agnostic representation learning}: We introduce \glsxtrlong{gl:dipa}, a novel method that trains models on balanced few-shot datasets using prior augmentations in order to make the model invariant to the class prior $p(y)$.
    \item \textbf{Enhanced regularization and robustness}: We demonstrate that our proposed method acts as an effective regularizer, stabilizing the training process and improving robustness on severely imbalanced target domains, in particular in low-shot regimes.
\end{enumerate}

\section{Related work}\label{sec:relatedwork}

\subsection{Few-shot learning paradigm}
Most supervised \gls{gl:ml} methods require large amounts of labeled data in order to achieve reasonable performance.
\glsxtrlong{gl:fsl}, on the contrary, deals with the scenario where only a very limited number of labeled samples is available.

\subsubsection{FSL as transfer vs. meta-learning}
A common framework in \gls{gl:fsl} is \gls{gl:mtl}.
In fact, both terms are often used interchangeably.
The core goal of \gls{gl:mtl} is to learn entire function spaces in order to quickly adapt to unseen, related tasks using only a few labeled samples.
Therefore, it is also commonly referred to as \emph{learning-to-learn}.
One of the most prominent \gls{gl:mtl} algorithms is \gls{gl:maml} and its variants \citep{Raghu19:ANIL,Tseng22:TIML}.

\emph{Transfer learning}, on the other hand, consists of training a model on a rather large set of labeled data before transferring it to a second, often unrelated, target task with subsequent fine-tuning. 
This concept has been widely used across various fields \citep{Vinija24:transferlearning,Alem22:transferlearningLCLU,Rouba23:heterogenoustransferlearning}.
\Citet{Chen18:fewshotclassification} provide a comprehensive evaluation of existing \gls{gl:fsl} approaches.
They find that traditional transfer learning achieves comparable or even superior performance on few-shot tasks compared to state-of-the-art \glsxtrlong{gl:mtl} approaches.

\subsubsection{FSL in remote sensing}
\Gls{gl:fsl} has been widely applied to the field of remote sensing \citep{Reuss25:MTLSSL,Tseng22:TIML, Russwurm20:MTL,Wang20:MTL,Tseng20:CropHarvest}.
\Citet{Tseng22:TIML} extended the concept of the original \gls{gl:maml} algorithm explicitly for agricultural monitoring by taking into account additional metadata such as the spatial coordinates.
\Citet{Reuss25:MTLSSL} provide a comprehensive cross-regional benchmark study using the few-shot crop-type dataset \EuroCropsML \citep{Reuss25:EML}.
Their findings show that, while \glsxtrlong{gl:mtl} achieves superior performances compared to regular transfer learning and self-supervised learning, it comes at the expense of increased computational costs.
Moreover, they highlight that none of the evaluated methods were capable of overcoming the discrepancy in distribution between the balanced train set and the imbalanced test set.

\subsubsection{Class imbalance and prior shift in FSL}

\Citet{Ochael23:classimbalanceFSL} provide a detailed evaluation and comparison of various existing \glsxtrlong{gl:fsl} methods under class imbalance.
They found that random oversampling during balanced training significantly improves performance and outperforms rebalancing loss functions, \eg the \gls{gl:fl}.

\paragraph{Prior distribution shift correct at inference-time}
Recent studies address the problem of prior distribution shifts often at inference time.
\Gls{gl:bbse} estimates the test distribution $p_\text{test}(y)$ to improve generalization for symptom-diagnose detection.
\citet{Kluger21:FPSA} directly tackles the problem of label (prior) and feature (covariate) distribution shift in few-shot crop-type classification using crop statistics, assuming that the distribution of the test set is known.
Specifically, to address the prior distribution shift, they reweigh the posterior probabilities.
\Citet{Sipka22:hitchiker} present a novel prior estimation approach based on confusion matrices.

\subsection{Dirichlet priors and distribution augmentation}
The \gls{gl:DirDist} is often used to model the prior in Bayesian statistics, \cf \cref{sec:methods}.
Among others, previous work addressed supervised clustering \citep{Daume05:DirichletClustering} and the utilization of Dirichlet priors within a Bayesian framework for regression \citep{Rademacher21:Dirichletpriors}.
The latter propose a Dirichlet prior because it possesses multiple desirable benefits:
\begin{description} 
    \item[Full support]
    The \gls{gl:DirDist} covers the full space of possible probability distributions.
    This means that the model can, technically, still learn the true data distribution, regardless of the accuracy of the initial prior.
    \item[Closed-form posterior distributions] 
    It represents the conjugate prior for multinomial data (\eg categorical counts), \cf \cref{sec:methods}.
    This leads to a simple, closed-form posterior distribution, which significantly simplifies mathematical derivations and provides computational efficiency.
    \item[Controllable informativeness]
    The prior contains a so-called localization parameter $\alpha_0$ which explicitly manages the bias-variance trade-off.
    It can be set to be highly opinionated (modeling strong prior knowledge) or non-committal (allowing the data to dominate the model's training).
\end{description}

\subsubsection{Dirichlet for FSL evaluation}

Although Dirichlet priors have been employed for few-shot learning, previous studies rely on the assumption that both the train (support) and test (query) sets are balanced.
Therefore, they utilize Dirichlet sampling to generate diverse test distributions \citep{Veilleux21:dirichletevaluation,Mohammadi24:fewshotcropmappingdirichlet}.
Thus, these approaches can be considered instantiating a few-shot evaluation method, since their sole effect is to simulate a realistic imbalanced test set.

\subsection{Summary and relation to our work}
While recent studies \citep{Reuss25:MTLSSL} demonstrate that \glsxtrlong{gl:mtl} methods often achieve slightly superior performance in \gls{gl:fsl} for crop-type classification, they suffer from high computational costs.
Therefore, this work chooses the transfer-learning paradigm, which has been shown to achieve competitive results \citep{Reuss25:MTLSSL,Chen18:fewshotclassification}.
However, the underlying principle of \gls{gl:dipa} is general and not restricted to this paradigm.

Addressing the problem of prior distribution shifts, existing methods primarily rely on correction at inference time \citep{Kluger21:FPSA,Sipka22:hitchiker}, requiring explicit or estimated knowledge of the final test distribution.
Moreover, while the Dirichlet distribution has been used in \gls{gl:fsl} in order to create diverse evaluation sets \citep{Veilleux21:dirichletevaluation,Mohammadi24:fewshotcropmappingdirichlet}, its utilization has been limited to evaluation only.

Our \glsxtrfull{gl:dipa} approach tackles this shift directly during training, representing a novel proactive approach.
By using the sampled priors to augment the training distribution with diverse class priors, \gls{gl:dipa} forces the model to learn a feature representation that is fundamentally \emph{prior-agnostic}, eliminating the need for any inference-time prior estimation.

\section{Dataset}\label{sec:dataset}
In this study, we use the Estonia data from the \EuroCropsML dataset \citep{Reuss25:EML} for training and evaluation.
\EuroCropsML is a time-series dataset that combines parcel reference data and multi-class \gls{gl:hcat} labels from \EuroCrops \citep{Schneider2023:EuroCrops} with Sentinel-2 L1C optical satellite observations captured during the year 2021.
Each data point contains a time series of cloud-free multi-spectral Sentinel-2 observations for all \num{13} bands.
We updated the original \EuroCropsML labels with the newest \gls{gl:hcat} version 4 \citep{Claverie25:ECM} to reflect the corrected class structure. 

The dataset reflects real-world agricultural complexity, including regional variations in crop types, vegetation patterns, and parcel sizes, which pose significant challenges for classification.
Notably, it also exhibits a strong class imbalance with \meadow being the most frequent one among the \num{129} crop types, representing \qty{46}{\percent} of all samples.

\subsection{Dataset split}
The total dataset comprising \num{175906} samples is divided into training, validation, and test sets.
We allocate 60\% of the samples to the training set and divide the remaining 40\% equally between the validation and testing sets.
This yields \num{105543} samples for training purposes and \num{35182} for both validation and testing.
The dataset's imbalanced distribution, where some classes contain only a single sample, created partly disjoint sets.
Specifically, \num{24} classes are unique to the training set.
In total, the test set contains \num{95} classes of which \num{7} are not present in the train set.
As a consequence, the model is forced to perform \emph{zero-shot} classification when attempting to classify any of these \num{7} novel classes.
Of these \num{7} classes, \num{6} are fully unique to test.

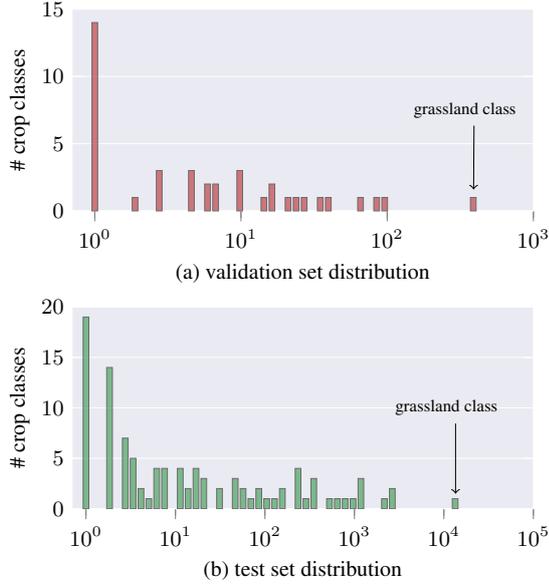
\begin{figure}[t]
    \centering
    \resizebox{.9\linewidth}{!}{\input{images/val_dist}}
    \hspace{.03\linewidth}
    \resizebox{.9\linewidth}{!}{\input{images/test_dist}}
    \caption{%
    Abundances of crop types in Estonia.
    Histograms showing the binned distribution of crop-type abundances in Estonia for \num{1000} randomly sampled data points of the validation set and the full test set.}
    \label{fig:dataset_dists}
\end{figure}

\begin{figure}[]
  \centering
  \includegraphics[width=\linewidth, trim={6cm 5cm 6cm 5cm}, clip]{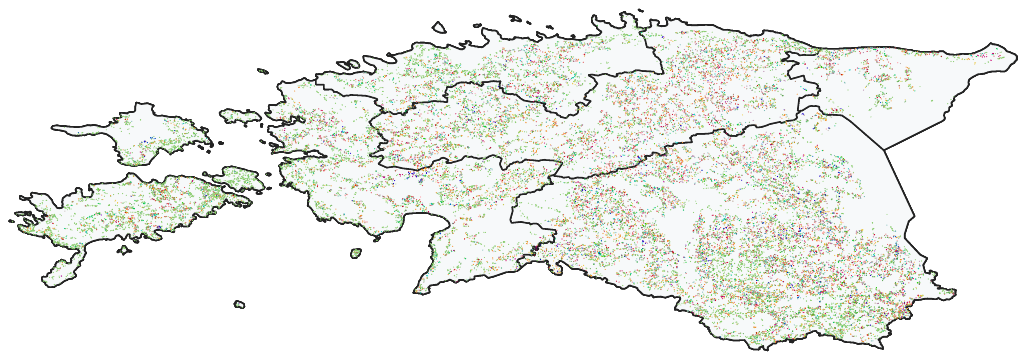}
  \caption{%
  Spatial distribution of crop classes in the test set of Estonia.
  Map of Estonia showing the location and distribution of crop types within the final test set. Each data point marks the central coordinate of an agricultural parcel and is color-coded by its corresponding crop class.
  }
  \label{fig:estonia_pixels}
\end{figure}

There some classes contain only a single sample---created partly disjoint sets.
\Cref{fig:dataset_dists} shows the distributions of crop-type abundances within the validation and test set, while \cref{fig:estonia_pixels} presents the spatial test class coverage.

We sample different few-shot scenarios, specifically:
\numlist[list-final-separator = {, or }]{1;5;10;20;100;200;500} shots.
Updating the crop classes to \gls{gl:hcat}4 gave rise to alterations to the classifications of certain parcels.
Therefore, the utilization of the original splits \citep{Reuss25:EML} would have resulted in a violation of the few-shot setting.

\section{Methodology}\label{sec:methods}
Our proposed method aims to simulate prior distributions shifts proactively during the training process to increase the robustness of the model against various potential label distributions during inference.
Instead of training with balanced priors, we inject synthetic prior shifts during training.

\subsection{Dirichlet distribution}
\label{sec:dirichlet}
The Dirichlet distribution, denoted $Dir(\boldsymbol{\alpha})$ and formally stated in \cref{def:dirichlet}, is a family of continuous multivariate probability distributions, parametrized by $\boldsymbol{\alpha} \in \mathbb{R}_+^K$.
It models the distribution of proportions or probabilities, \ie non-negative values with unit integral, and is commonly used as a prior in Bayesian statistics \citep{Harald02:DirichletBayes,Daume05:DirichletClustering,Rademacher21:Dirichletpriors}.

\begin{definition}
    \label{def:dirichlet}
    Let $\boldsymbol{X} = (X_1, \ldots, X_K) \in \mathbb{R}^K$ be a $K$-dimensional continuous random vector.
    The Dirichlet distribution is defined for $K \geq 2$ variables and parameterized by the $K$-dimensional concentration parameter vector $\boldsymbol{\alpha} = (\alpha_1, \ldots, \alpha_K), \alpha_c > 0 \ \forall c \in \{1,\ldots,K\}$.
    The probability density function of $\boldsymbol{X}$ is given by
    \begin{align}
    p(\boldsymbol{x} \mid \boldsymbol{\alpha}) = \frac{1}{B(\boldsymbol{\alpha})} \prod_{c=1}^{K} x_c^{\alpha_c-1} \quad,
    \end{align}
    where $x_c \in \left[0,1\right] \ \forall c \in \{1,\ldots,K\}$ with $\sum\limits_{c=1}^K x_c = 1$ and $B(\boldsymbol{\alpha}) = \frac{\prod_{c=1}^K \Gamma(\alpha_c)}{\Gamma \left( \sum_{c=1}^K \alpha_c \right)}$ being the multivariate Beta function which can be expressed using the \emph{Gamma} function $\Gamma$.\\
    The symmetric form of the Dirichlet distribution implies no prior knowledge of $p_c$, \ie $\mathop{\mathbb{E}}[p_c] = \frac{1}{K}$.
    It is denoted as $Dir(\alpha \cdot \boldsymbol{1})$.
\end{definition}

\begin{figure*}[t]
  \centering
  \subcaptionbox{$\boldsymbol{\alpha}=(0.5, 0.5, 0.5)$ \label{fig:dir_a05}}[.24\linewidth]{\includegraphics[width=\linewidth, keepaspectratio]{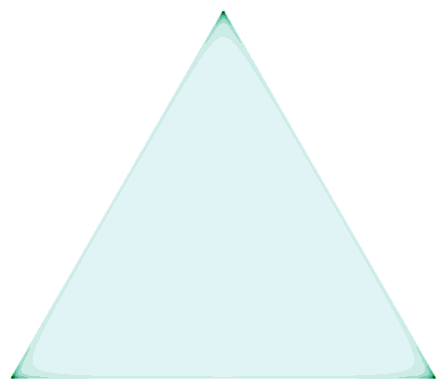}}
  \hspace{1em}
  \subcaptionbox{$\boldsymbol{\alpha}=(30, 30, 30)$ \label{fig:dir_a30}}[.24\linewidth]{\includegraphics[width=\linewidth, keepaspectratio]{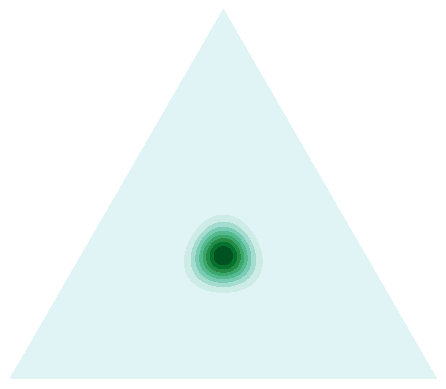}}
  \hspace{1em}
  \subcaptionbox{$\boldsymbol{\alpha}=(5, 5, 5)$ \label{fig:dir_a5}}[.24\linewidth]{\includegraphics[width=\linewidth, keepaspectratio]{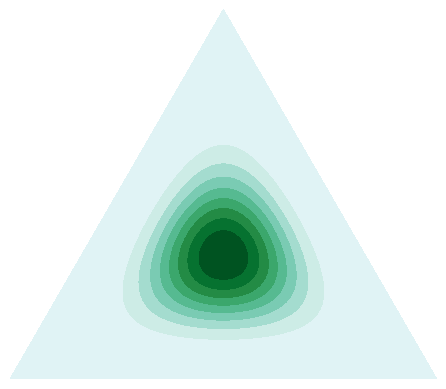}}
  \hspace{1em}
  \subcaptionbox*{\label{fig:dir_cbar}}[.05\linewidth]{%
    \resizebox{!}{3.5cm}{\input{images/dirichlet_densitycolorbar}}
  }\hspace*{0.04\linewidth}
  \caption{Dirichlet density for $K=3$ (defined over the $(K-1)=2$-simplex) and different concentration parameters $\boldsymbol{\alpha}$.}
  \label{fig:dirichlet_density}
\end{figure*}

\Cref{fig:dirichlet_density} illustrates the density function of the (symmetric) Dirichlet distribution for $K=3$ variables and different $\boldsymbol{\alpha}$.

\subsection{DiPA: Dirichlet prior augmentation}
To illustrate the main idea, consider a labeled dataset $\mathcal{D} = \{(\boldsymbol{x}_i,y_i)\}_{i=1}^n$ of multivariate time series, where each $\boldsymbol{x}_i \in \mathbb{R}^{n_t\times d}$ has $n_t \leq T_\text{max}$ time steps and $d$ channels, and $y_i \in \mathcal{C} = \{1, \ldots,K\}$ is the corresponding class label.
We denote $\boldsymbol{z}_i = f_{\boldsymbol{\theta}}(\boldsymbol{x}_i)$, where $f_{\boldsymbol{\theta}} \colon \mathbb{R}^{T_\text{max}\times d} \to \mathbb{R}^K$ is a model parameterized by $\boldsymbol{\theta}$ that maps observations $\boldsymbol{x}_i$ to a vector of logits $\boldsymbol{z}_i \in \mathbb{R}^K$.
Subsequently, in general multi-class classification problems, the predictive distribution 
\begin{align}
\hat{\boldsymbol{p}}_i = \sigma(\boldsymbol{z}_i) = \sigma(f_{\boldsymbol{\theta}}(\boldsymbol{x}_i))
\end{align}
of $K$ classes is obtained by applying the \emph{Softmax} function $\sigma$ to the logits.

In order to augment the training distribution and make the model more robust against unknown test skews, at each training step $s = 1, \ldots, S$, we introduce a class prior $\tilde{\boldsymbol{\pi}}^{(s)}$ to the model's output logits via the following steps:

\begin{enumerate}
    \item \textbf{Sample pseudo-prior}: 
    We sample a pseudo-prior vector 
    \begin{align}
      \tilde{\boldsymbol{\pi}}^{(s)} = (\tilde{\pi}_1^{(s)}, \ldots, \tilde{\pi}_K^{(s)})
      &\sim Dir(\alpha \cdot \boldsymbol{1})
    \end{align}
    from the symmetric Dirichlet distribution $Dir(\alpha \cdot \boldsymbol{1})$. 
    The parameter $\alpha \in \mathbb{R}^+$ controls the degree of imbalance, with $\alpha < 1$ sampling highly skewed (imbalanced) distributions, and $\alpha > 1$ sampling distributions closer to uniform.
    \item \textbf{Logit adjustment}: 
    We use the sampled prior and a scaling factor $\tau \in \mathbb{R}^+$ to adjust the logits $\boldsymbol{z}_i$.
    The adjusted logits 
    \begin{align}
    \boldsymbol{z}_i^\prime &\leftarrow \boldsymbol{z}_i + \tau \log (\tilde{\boldsymbol{\pi}}^{(s)})
    \end{align}
    are computed element-wise.
\end{enumerate}

Subsequently, the predictive probability distribution $\hat{\boldsymbol{p}}_i = \sigma(\boldsymbol{z}_i^\prime)$ is computed by applying the \emph{Softmax} function to the adjusted logits. 
Since we assume that we have no prior knowledge about the actual test distribution, we sample from the \emph{symmetric} Dirichlet distribution.
The full pseudo-code is outlined in \cref{alg:symm_dipa}.

By applying \gls{gl:dipa}, the model sees many possible class-frequency scenarios and learns a representation that works under varied priors.

\begin{algorithm}
\caption{Dirichlet prior augmentation}\label{alg:symm_dipa}
\begin{algorithmic}[1]
\Require $\alpha, \tau \in \mathbb{R}^+$
\Require $f_{\boldsymbol{\theta}}$ (model parameterized by $\boldsymbol{\theta}$)
\For{each training step $s = 1$ to $S$}
\State sample mini-batch of data points $D^{(s)} = \{(\boldsymbol{x}_i, y_i)\}_{i=1}^{b}$
\State sample pseudo-prior $\tilde{\boldsymbol{\pi}}^{(s)} \sim Dir(\alpha \cdot \boldsymbol{1})$
\For{each data point $(\boldsymbol{x}_i, y_i)$ in $D^{(s)}$}
    \State compute base logits: $\boldsymbol{z}_i \leftarrow f_{\boldsymbol{\theta}}(\boldsymbol{x}_i)$
    \State adjust logits: $\boldsymbol{z}_i' \leftarrow \boldsymbol{z}_i + \tau \log (\tilde{\boldsymbol{\pi}}^{(s)})$
    \State compute predictive distribution: $\hat{\boldsymbol{p}}_i \leftarrow \sigma(\boldsymbol{z}_i')$
\EndFor
\State compute mini-batch loss \\
$\mathcal{L}_{\text{batch}} \leftarrow \frac{1}{b}\sum_{i=1}^{b} \mathcal{L}(\hat{\boldsymbol{p}}_i, y_i)$
\EndFor
\end{algorithmic}
\end{algorithm}

\subsection{Transformer model}
\label{sec:transformer}
All experiments, as described in \cref{sec:experiments}, are conducted using a state-of-the-art Transformer encoder architecture with sinusoidal positional encoding \citep{vaswani_vanillatransformer,Schneider21:SIT}. 
We set the maximum sequence length $T_\text{max}$ to \num{366} days, \ie a full year, including one leap day. 
This encoder serves as the model's feature extractor, which we call the \emph{backbone} and denote it as 
\begin{align}
  f^\text{backbone}_{\boldsymbol{\theta}_\text{backbone}} &\colon \mathbb{R}^{T_\text{max} \times d}\to\mathbb{R}^{n_e},
\end{align}
where $n_e \in \mathbb{N}$ denotes the Transformer embedding dimension and $\boldsymbol{\theta}_\text{backbone}$ all trainable model parameters of the backbone.
The encoder's output is subsequently fed into a single linear layer to map the extracted features to the final class logits.
We refer to this classification layer as the \emph{head} of the model and denote it as
\begin{align}
  f^\text{head}_{\boldsymbol{\theta}_\text{head}} &\colon \mathbb{R}^{n_e} \to \mathbb{R}^{K},
\end{align}
As before, $\boldsymbol{\theta}_\text{head}$ collects all of the head's trainable parameters.

The complete end-to-end model is given by the composition
\begin{align*}
  f_{\boldsymbol{\theta}} &= f^\text{head}_{\boldsymbol{\theta}_\text{head}} \circ f^\text{backbone}_{\boldsymbol{\theta}_\text{backbone}},
\end{align*}
where $\boldsymbol{\theta}=\left[\boldsymbol{\theta}_\text{backbone},\boldsymbol{\theta}_\text{head}\right]$ represents the full set of all trainable model parameters.

\section{Experiments}\label{sec:experiments}
In all experiments, we use a single Transformer encoder block with four attention heads.
Each token in the input sequence is represented by an internal embedding vector of dimension \num{128}.
This is further expanded by the fully connected network within the Transformer block, which employs a hidden dimension of $d_\text{hidden}=256$.
We apply additive sinusoidal temporal positional encoding \citep{vaswani_vanillatransformer} with a maximum sequence length of \num{366}, encompassing daily samples over the span of a full year, including leap years, \cf \cref{sec:transformer}.%
In order to predict the class log-probabilities, we apply a linear classification layer on top.

We train a randomly initialized network from scratch and fine-tune a model pretrained on the \EuroCropsML Latvian data \citep{Reuss25:MTLSSL}.
For the pretrained model, we reset the classification head $f^\text{head}$ and reinitialize it with the \num{129} classes from our target training set.
All models are trained end-to-end for up to \num{200} epochs with a batch size of \num{16} following a standard training paradigm with train, validation, and test sets.
Hyperparameters, such as the learning rate $\beta$, the focal focusing parameter $\gamma$, and the Dirichlet parameters $\alpha$ and $\tau$, are optimized on the \num{1000} fixed validation data points.
If the validation loss does not improve for \num{15} epochs, we stop training.
The final models are evaluated on all \num{35224} test samples, containing \num{102} unique classes.
All experiments are repeated five times, each time with a different random seed $r \in \{0, 1, 42, 123, 1234\}$ in order to evaluate the robustness of the results.

To assess the impact of the prior adjustment across different few-shot scenarios, we train the models in various few-shot settings with 
\numlist[list-final-separator = {, or }]{1;5;10;20;100;200;500} shots
\citep[\cf][]{Reuss25:EML}.
It is important to note that in our specific few-shot learning regime, tasks are sampled from the originally imbalanced training set.
Consequently, the actual number of samples available per class is constrained by the underlying label distribution.
Thus, as the target shot count increases, the number of samples per class can often be limited.
This, in turn, results in the \gls{gl:fsl} prior $p_\text{train}(y)$ converging towards the true empirical prior of the full training set.
Hence, the few-shot settings presented in this work test the method's generalization performance regarding both the balanced and empirical underlying training label distributions. 
In addition, we train the models on the entire Estonian training data to establish a baseline for the task's complexity in a standard (non-few-shot) setting.
We refer to this setting with \textit{all}.

We conduct all experiments with two different loss functions---\ie \gls{gl:ce} and \gls{gl:fl} (without a class-imbalance factor)---in order to evaluate the effect of the prior augmentation across various settings.
Furthermore, using \gls{gl:fl} allows for analyzing the impact of \gls{gl:dipa} in combination with a class-agnostic and difficulty-aware loss function. 
Both loss functions are trained with and without Dirichlet priors.

\section{Results and discussion}\label{sec:results}
We choose the overall classification accuracy as our core validation metric.
However, when working with highly imbalanced data, accuracy is often biased towards the majority class.
Therefore, we also report \emph{Cohen's kappa} ($\kappa$) as an additional evaluation metric for the final models.
We always report the test metrics of the best-performing models, measured in terms of validation accuracy.

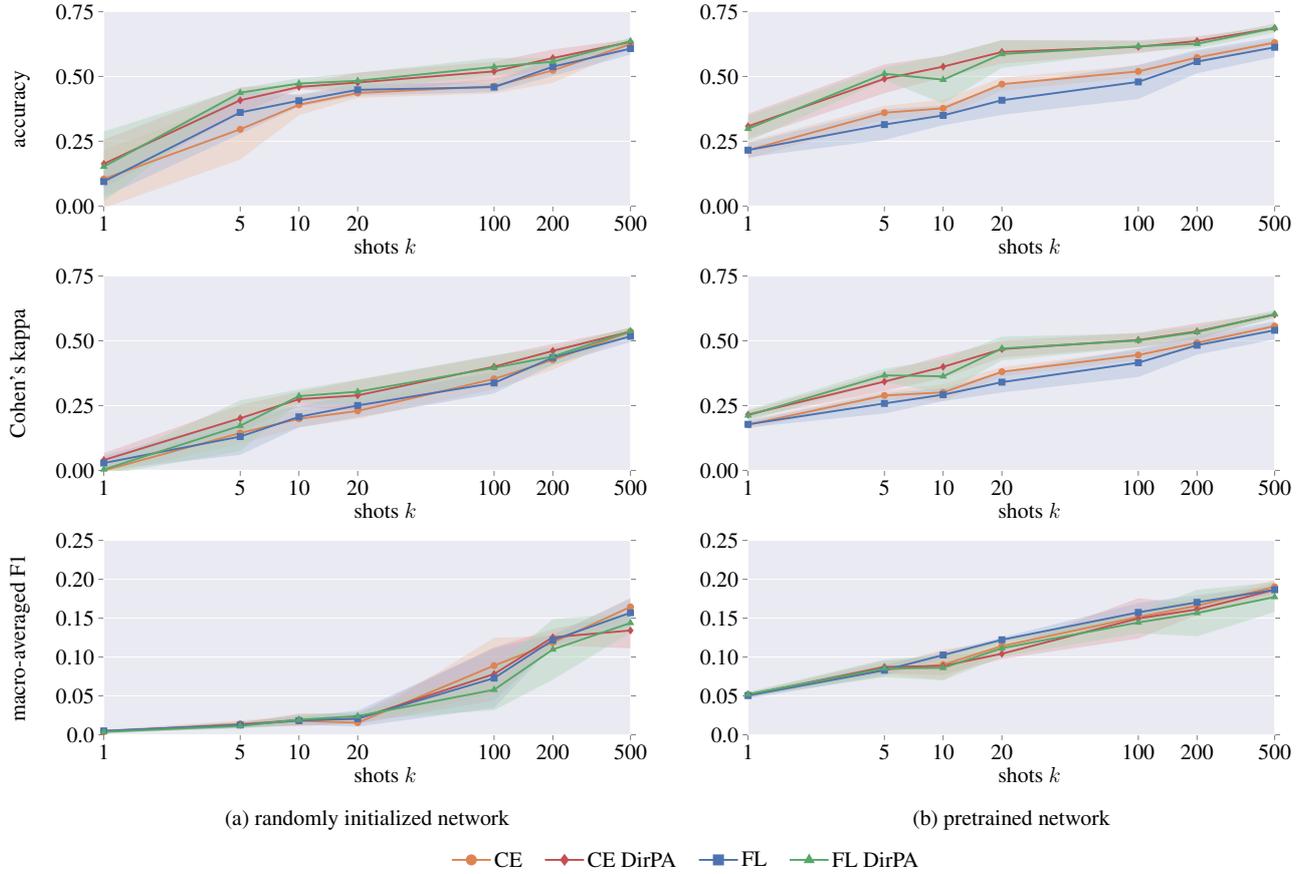
\begin{figure*}[t!]
    \resizebox{\linewidth}{!}{\input{images/results_plot_plusf1}}
     \caption{%
     Visualization of test metrics (including macro-averaged F1 score) across the $k$-shot benchmark tasks.
    The $x$-axis is plotted on a logarithmic scale.
    Metrics are shown as mean $\pm$ standard deviation over five runs,
    \cf \cref{sec:experiments}.
    The postfix \gls{gl:dipa} indicates the use of the \gls{gl:dipa} method.
    Due to the highly imbalanced nature of the \num{102}-class classification task, the macro-F1 scores remain numerically low, reflecting the inherent challenge in achieving high performance on the numerous low-resource classes.
    }
    \label{fig:results}
\end{figure*}

The results for the randomly initialized model are shown in \cref{tab:results_random} whilst those for the pretrained one are displayed in \cref{tab:results_pretrain}.
A graphical visualization of the results is provided in \cref{fig:results}.
Results are reported using the two aforementioned loss functions: \gls{gl:ce} and \gls{gl:fl}.
The postfix \gls{gl:dipa} is appended to the name of the loss function if the \gls{gl:dipa} method has been utilized.

\begin{description}
    \item[Randomly initialized model] 
    For the randomly initialized model, across all few-shot settings, \gls{gl:dipa} improved or matched the baseline.
    For \gls{gl:ce}, adding prior augmentation achieved higher overall accuracy and kappa scores across all few-shot tasks from \num{1}- to \num{200}-shot, with the largest gains observed in the low-shot regime ($k \leq 20$). 
    The improvement remained up to the \num{500} setting, where both variants converged to nearly identical results.
    For \gls{gl:fl}, \gls{gl:dipa} yielded higher scores in both metrics for nearly all few-shot configurations, except for \num{1} and \num{200} samples.
    The \gls{gl:ce}-based models achieved better performance in the \num{1}-shot and \num{200}-shot scenarios, while \gls{gl:fl}-based ones overtake \num{5}-, \num{10}-, and \num{20}-shot. 
    \item[Pretrained model] 
    When fine-tuning the model pretrained on Latvian data, \gls{gl:dipa} improved both metrics in all few-shot regimes. 
    \gls{gl:ce} \gls{gl:dipa} achieved higher accuracy and Cohen’s kappa, compared to \gls{gl:ce}, for every few-shot task, with differences shrinking in the \textit{all} setting.
    For \gls{gl:fl}, the \gls{gl:dipa} setting also surpassed the baseline across all settings.
    \gls{gl:fl} \gls{gl:dipa} obtained the highest accuracy among all models in most few-shot settings, while \gls{gl:ce} \gls{gl:dipa} was slightly better at 
    \numlist[list-final-separator = {, or }]{5;200;500} shots. 
    The difference between \gls{gl:ce} \gls{gl:dipa} and \gls{gl:fl} \gls{gl:dipa} narrowed as shots increased, and all variants exhibited similar performance in the \textit{all} setting.
\end{description}

Across both initialization regimes, the Dirichlet prior consistently improves both accuracy and Cohen’s kappa, confirming \gls{gl:dipa} benefits model robustness under label imbalance. 
The largest relative advantage appears in the low-shot regimes.
By dynamically sampling a skewed pseudo prior vector $\tilde{\pi}^{(s)}$ at every step, the method serves as a feature regularizer, especially in low-shot regimes.
This forces the model to stabilize predictions where data is sparse.

The results also indicate that \gls{gl:dipa} does not degrade performance when the number of available samples increases. 
In higher-shot or full-data conditions, all methods converge to nearly identical results.
This almost identical performance is as expected since the training prior converges towards the empirical prior of the full Estonia data, dissolving the positive effect of the Dirichlet priors.
The improved kappa scores indicate that the model's predictions are more robust and less likely due to chance.
While initial experiments suggest that macro metrics show inferior performance for the \gls{gl:dipa} method (\cf \cref{fig:results}), this constitutes a necessary trade-off for achieving superior overall performance and stability:
The concentrated loss often occurs on stable, high-shot classes, where \gls{gl:dipa}'s strong dynamic regularization over-smoothed already established decision boundaries.
However, the consistent gain in overall system reliability ($\kappa$) justifies this small, concentrated loss on single classes, as the final model is demonstrably more robust for classifying the total volume of crops.

\section{Conclusions and future work}\label{sec:conclusion}
This study proposed \glsxtrfull{gl:dipa}, a novel method designed to bridge the gap between training and test priors in real-world few-shot crop-type classification.
\Gls{gl:dipa} augments the balanced few-shot training data with dynamically sampled pseudo-priors from the Dirichlet distribution.
This process acts as a robust regularizer, improving generalization and stability of the model across imbalanced test data.
It is applied directly during the training process and does not require any knowledge about the final test distribution.
We evaluated the method against the challenging task of classifying \num{102} heterogeneous, highly imbalanced crop types in Estonia.
The evaluation involved two distinct loss functions, namely \gls{gl:ce} and \gls{gl:fl}.
We demonstrated that \gls{gl:dipa} improved overall accuracy and Cohen's kappa across various few-shot regimes.
Although this study focused on crop-type classification, \gls{gl:dipa} can be applied to any few-shot learning task that suffers from a discrepancy between the training and test label distributions.

Future work will investigate applying pseudo-priors sampled from an asymmetric Dirichlet distribution while still assuming an unknown but imbalanced test prior, as well as extensive hyperparameter tuning.
Moreover, we will test the efficacy of our method on additional countries of the European Union and investigate the potential of \gls{gl:dipa} with regard to enhancing not only the system's stability but also class-specific performance metrics.

\begin{table*}[ht!]
    \centering
    \scriptsize
    \begin{subtable}{\linewidth}
        \centering
        \input{tables/results_random} 
        \caption{%
        randomly initialized network
        }
        \label{tab:results_random}
        \vspace{10pt}
    \end{subtable}%
    \hfill
    \begin{subtable}{\linewidth}
        \centering
        \input{tables/results_pretrain} 
        \caption{%
         pretrained network
        }
        \label{tab:results_pretrain}      
    \end{subtable}
    \caption{%
    Few-shot classification results.
    We report the \textit{classification accuracy} and \textit{Cohen's kappa} on the test set for each variant and few-shot task.
    Test metrics are shown as mean $\pm$ standard deviation over five runs, \cf \cref{sec:experiments}. 
    The best result for each few-shot scenario is highlighted in {\color{blue}\textbf{bold blue}}.
    Depending on which loss function achieves the best result, the top result of the other loss function is shown in \textbf{bold black}. 
    The postfix \gls{gl:dipa} indicates the use of the \gls{gl:dipa} method.
    }     
    \vspace{15pt}
\end{table*}
\hfill

\clearpage
\twocolumn[\null]
\section*{Author contributions statement}
J{.} Reuss planned the study conception and design and was responsible for the implementation of the methodology, as well as for the experimental setup, the conduction of experiments, and their investigation and validation.
J{.} Reuss and E{.} Gikalo collected and validated the data.
J{.} Reuss and E{.} Gikalo performed the formal analysis of the experiments and created the visualizations.
J{.} Reuss, E{.} Gikalo, and M{.} K\"orner contributed to the writing of the article.
M{.} K\"orner was responsible for the funding acquisition and supervised the project.
All authors contributed to the article, reviewed the manuscript, and approved the submitted version.

\section*{Acknowledgments}\label{sec:acknowledgements}
The project is funded by the German Federal Ministry for Economic Affairs and Energy based on a decision by the German Bundestag under the funding references 50EE2007B (J{.} Reuss, E{.} Gikalo, and M{.} K{\"o}rner) and 50EE2105 (M{.} K{\"o}rner).

\footnotesize
\bibliography{main}

\end{document}

%% file: acronyms.tex
\newacronym{gl:dipa}{DirPA}{\textbf{Dir}ichlet \textbf{P}rior \textbf{A}ugmentation}

\newacronym{gl:ai}{AI}{artificial intelligence}
\newacronym{gl:dl}{DL}{deep learning}
\newacronym{gl:ml}{ML}{machine learning}
\newacronym{gl:fsl}{FSL}{few-shot learning}
\newacronym{gl:mtl}{MTL}{meta-learning}
\newacronym{gl:nn}{NN}{neural network}
\newacronym[user1={Finn17:MAML}]{gl:maml}{MAML}{model-agnostic meta-learning}
\newacronym[user1={Raghu19:ANIL}]{gl:anil}{ANIL}{almost no inner loop}
\newacronym[user1={Tseng22:TIML}]{gl:timl}{TIML}{task-informed meta-learning}
\newacronym[user1={Lipton18:BlackBox}]{gl:bbse}{BBSE}{Black Box Shift Estimation}

\newacronym{gl:vit}{ViT}{Vision Transformer}
\newacronym{gl:cv}{CV}{computer vision}
\newacronym{gl:mlp}{MLP}{multi-layer perceptron}
\newacronym[user1={Lin18:FocalLoss}]{gl:fl}{FL}{focal loss}
\newacronym{gl:ce}{CE}{cross-entropy loss}

\newacronym{gl:eu}{EU}{European Union}
\newacronym{gl:esa}{ESA}{European Space Agency}
\newacronym{gl:gee}{GEE}{Google Earth Engine}

\newacronym{gl:mse}{MSE}{mean squared error}
\newacronym{gl:tpe}{TPE}{bayesian tree-parzen estimator}
\newacronym{gl:sgd}{SGD}{stochastic gradient descent}

\newacronym{gl:cap}{CAP}{common agriculture policy}
\newacronym{gl:nuts}{NUTS}{Nomenclature of Territorial Units for Statistics}

\newacronym[user1={Schneider2023:EuroCrops,Schneider2021:TinyEuroCrops}]{gl:hcat}{HCAT}{hierarchical crop and agriculture taxonomy}

\newacronym{gl:eo}{EO}{Earth observation}

\newglossaryentry{gl:DirDist}{name={Dirichlet distribution}, description={}}

%% file: images/val_dist.tex
\begin{tikzpicture}

\definecolor{purple}{RGB}{129, 114, 179}
\definecolor{indianred}{RGB}{196,78,82}
\definecolor{lavender}{RGB}{234,234,242}
\definecolor{mediumseagreen}{RGB}{85,168,104}
\definecolor{peru}{RGB}{221,132,82}
\definecolor{steelblue}{RGB}{76,114,176}

\pgfplotstableread[col sep=comma]{images/val_binned.csv}\table

\pgfplotsset{
BarPlot/.style={
    axis background/.style={fill=lavender},
    axis line style={white},
    font=\small,
    ybar stacked,     
    bar width=0.01\linewidth,
    width=\linewidth,
    height=4.5cm,
    ymin=0.0,
    ymax=15,
    ytick={0,5,10,15},
    ymajorgrids,
    ymajorticks=true,
    ytick style={color=white},
    y grid style={white},
    xmin=0.7, 
    xmax=1000,
    minor xtick={1,10,100,1000},
    xmajorticks=false,
    xtick pos=left,
    xtick align=outside,
    xtick=false,
}
}

\begin{axis}[BarPlot,
    xmode=log,
    log base x=10,
    xmajorticks=true,
    ylabel=\# crop classes,
    ylabel style={
    at={(axis description cs:-0.075,0.5)}, 
    anchor=south,                        
    },
    xlabel=(a) validation set distribution,
]
\addplot[draw=darkgray, fill=indianred, opacity=0.75]
    table[x=bin_start, y=num_classes]{\table};
    \node[anchor=east] (meadowEEtext) at (axis description cs:0.98,0.5) {\scriptsize grassland class};
\coordinate (meadowEE) at (axis cs:390,1.6);
\end{axis}

\draw[->] (meadowEEtext.300) -- (meadowEE);

\end{tikzpicture}

%% file: images/test_dist.tex
\begin{tikzpicture}

\definecolor{purple}{RGB}{129, 114, 179}
\definecolor{indianred}{RGB}{196,78,82}
\definecolor{lavender}{RGB}{234,234,242}
\definecolor{mediumseagreen}{RGB}{85,168,104}
\definecolor{peru}{RGB}{221,132,82}
\definecolor{steelblue}{RGB}{76,114,176}

\pgfplotstableread[col sep=comma]{images/test_binned.csv}\table

\pgfplotsset{
BarPlot/.style={
    axis background/.style={fill=lavender},
    axis line style={white},
    font=\small,
    ybar stacked,     
    bar width=0.01\linewidth,
    width=\linewidth,
    height=4.5cm,
    ymin=0.0,
    ymax=20,
    ytick={0,5,10,15,20},
    ymajorgrids,
    ymajorticks=true,
    ytick style={color=white},
    y grid style={white},
    xmin=0.7, 
    xmax=100000,
    minor xtick={1,10,100,1000,10000,100000},
    xmajorticks=false,
    xtick pos=left,
    xtick align=outside,
    xtick=false,
}
}

\begin{axis}[BarPlot,
    xmode=log,
    log base x=10,
    xmajorticks=true,
    ylabel=\# crop classes,
    ylabel style={
    at={(axis description cs:-0.075,0.5)}, 
    anchor=south,                        
    },
    xlabel= (b) test set distribution,
]
\addplot[draw=darkgray, fill=mediumseagreen, opacity=0.75]
    table[x=bin_start, y=num_classes]{\table};
    \node[anchor=east] (meadowEEtext) at (axis description cs:0.94,0.5) {\scriptsize grassland class};
\coordinate (meadowEE) at (axis cs:13500,1.5);
\end{axis}

\draw[->] (meadowEEtext.300) -- (meadowEE);

\end{tikzpicture}

%% file: images/dirichlet_densitycolorbar.tex
\begin{tikzpicture}

\node (img1) at (0,0) {\includegraphics[height=3.5cm, width=0.3cm]{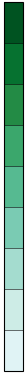}};

\node[anchor=south] at ($(img1.north) + (0, 0.05)$) (upperlimittest) {\large high density};

\node[anchor=north] at ($(img1.south) + (0, -0.05)$) (lowerlimittest) {\large low density};

\end{tikzpicture}

%% file: tables/results_random.tex
\robustify\bfseries
\sisetup{detect-all=true,uncertainty-mode=separate,table-align-uncertainty=true,round-mode=uncertainty,round-precision=2}



\begin{tabular}{llSSSSSSSS}
\toprule
& algorithm & \multicolumn{7}{l@{}}{benchmark task ($k$-shot)} \\
\cmidrule(l){3-10}
{} & {} & {1} & {5} & {10} & {20} & {100} & {200} & {500} & {all}\\
\cmidrule(lr){1-2} \cmidrule(lr){3-3} \cmidrule(lr){4-4} \cmidrule(lr){5-5} \cmidrule(lr){6-6} \cmidrule(lr){7-7} \cmidrule(lr){8-8} \cmidrule(l){9-9} \cmidrule(l){10-10}

\multirow[c]{4}{*}{\rotatebox[origin=c]{90}{accuracy}} & \gls{gl:ce} & \num{0.104548 +- 0.112009} & \num{0.296168 +- 0.116567} & \num{0.390472 +- 0.040536} & \num{0.436035 +- 0.018517} & \num{0.462037 +- 0.027566} & \num{0.523438 +- 0.048271} & \num{0.624342 +- 0.018480} & \num{0.783344 +- 0.002905} \\
 & \gls{gl:ce} \gls{gl:dipa} & \colorblue \bfseries \num{0.162759 +- 0.093704} & \bfseries \num{0.408078 +- 0.050266} & \bfseries \num{0.459428 +- 0.009415} & \bfseries \num{0.476568 +- 0.038176} & \bfseries \num{0.519555 +- 0.037690} & \colorblue \bfseries \num{0.570883 +- 0.033272} & \bfseries \num{0.633176 +- 0.010484} & \colorblue \bfseries \num{0.784162 +- 0.003388} \\
 & \gls{gl:fl}  & \num{0.094401 +- 0.059841} & \num{0.360991 +- 0.082588} & \num{0.406441 +- 0.022837} & \num{0.448349 +- 0.022247} & \num{0.458581 +- 0.018549} & \num{0.537570 +- 0.035559} & \num{0.607481 +- 0.023350} & \num{0.776983 +- 0.005830} \\
 & \gls{gl:fl} \gls{gl:dipa} & \bfseries \num{0.152163 +- 0.135363} & \colorblue \bfseries \num{0.437161 +- 0.017142} & \colorblue \bfseries \num{0.472890 +- 0.020433} & \colorblue \bfseries \num{0.483054 +- 0.030980} & \colorblue \bfseries \num{0.536735 +- 0.034560} & \bfseries \num{0.556813 +- 0.013951} & \colorblue \bfseries \num{0.636070 +- 0.013383} & \bfseries \num{0.778915 +- 0.002268} \\

\cmidrule(rl){1-10}

\multirow[c]{4}{*}{\rotatebox[origin=c]{90}{kappa}} & \gls{gl:ce} & \num{-0.000338 +- 0.016347} & \num{0.144304 +- 0.043103} & \num{0.199047 +- 0.032757} & \num{0.229330 +- 0.030665} & \num{0.353096 +- 0.040843} & \num{0.426336 +- 0.037357} & \num{0.535237 +- 0.016005} & \num{0.708631 +- 0.003473} \\
 & \gls{gl:ce} \gls{gl:dipa} & \colorblue \bfseries \num{0.039949 +- 0.026687} & \colorblue \bfseries \num{0.201272 +- 0.051831} & \bfseries \num{0.274613 +- 0.029008} & \bfseries \num{0.289456 +- 0.058642} & \colorblue \bfseries \num{0.400292 +- 0.041715} & \colorblue \bfseries \num{0.461172 +- 0.026410} & \bfseries \num{0.535732 +- 0.012809} & \colorblue \bfseries \num{0.710070 +- 0.003609} \\
 & \gls{gl:fl}  & \bfseries \num{0.028380 +- 0.028068} & \num{0.130522 +- 0.070531} & \num{0.206744 +- 0.041000} & \num{0.250533 +- 0.045621} & \num{0.337405 +- 0.040373} & \num{0.435228 +- 0.028094} & \num{0.517345 +- 0.022225} & \num{0.701026 +- 0.006203} \\
 & \gls{gl:fl} \gls{gl:dipa} & \num{0.003640 +- 0.022660} & \bfseries \num{0.172151 +- 0.097872} & \colorblue \bfseries \num{0.286557 +- 0.025967} & \colorblue \bfseries \num{0.303693 +- 0.047307} & \bfseries \num{0.395811 +- 0.048796} & \bfseries \num{0.439489 +- 0.034143} & \colorblue \bfseries \num{0.537616 +- 0.014995} & \bfseries \num{0.701875 +- 0.002039} \\
\bottomrule
\end{tabular}

%% file: tables/results_pretrain.tex
\robustify\bfseries
\sisetup{detect-all=true,uncertainty-mode=separate,table-align-uncertainty=true,round-mode=uncertainty,round-precision=2}



\begin{tabular}{llSSSSSSSS}
\toprule
& algorithm & \multicolumn{7}{l@{}}{benchmark task ($k$-shot)} \\
\cmidrule(l){3-10}
{} & {} & {1} & {5} & {10} & {20} & {100} & {200} & {500} & {all}\\
\cmidrule(lr){1-2} \cmidrule(lr){3-3} \cmidrule(lr){4-4} \cmidrule(lr){5-5} \cmidrule(lr){6-6} \cmidrule(lr){7-7} \cmidrule(lr){8-8} \cmidrule(l){9-9} \cmidrule(l){10-10}

\multirow[c]{4}{*}{\rotatebox[origin=c]{90}{accuracy}} & \gls{gl:ce} & \num{0.215343 +- 0.033730} & \num{0.360076 +- 0.026307} & \num{0.377062 +- 0.033261} & \num{0.470206 +- 0.024917} & \num{0.519465 +- 0.022799} & \num{0.572793 +- 0.022838} & \num{0.630698 +- 0.009358} & \num{0.789864 +- 0.004101} \\
 & \gls{gl:ce} \gls{gl:dipa} & \colorblue \bfseries \num{0.307720 +- 0.048449} & \bfseries \num{0.490859 +- 0.055736} & \colorblue \bfseries \num{0.537951 +- 0.041593} & \colorblue \bfseries \num{0.594463 +- 0.045027} & \bfseries \num{0.614337 +- 0.024664} & \colorblue \bfseries \num{0.636718 +- 0.018197} & \colorblue \bfseries \num{0.686777 +- 0.007904} & \colorblue \bfseries \num{0.791365 +- 0.002331} \\
 & \gls{gl:fl}  & \num{0.215690 +- 0.027543} & \num{0.314291 +- 0.058924} & \num{0.349650 +- 0.038026} & \num{0.408158 +- 0.056310} & \num{0.478972 +- 0.065889} & \num{0.557296 +- 0.045052} & \num{0.612450 +- 0.038258} & \num{0.784583 +- 0.002840} \\
 & \gls{gl:fl} \gls{gl:dipa} & \bfseries \num{0.298164 +- 0.049403} & \colorblue \bfseries \num{0.510414 +- 0.024598} & \bfseries \num{0.487369 +- 0.091749} & \bfseries \num{0.586499 +- 0.053673} & \colorblue \bfseries \num{0.616412 +- 0.020125} & \bfseries \num{0.626099 +- 0.018064} & \bfseries \num{0.686408 +- 0.017954} & \bfseries \num{0.788250 +- 0.004721} \\

\cmidrule(rl){1-10}

\multirow[c]{4}{*}{\rotatebox[origin=c]{90}{kappa}} & \gls{gl:ce} & \num{0.176427 +- 0.016872} & \num{0.289047 +- 0.016864} & \num{0.301029 +- 0.032194} & \num{0.380613 +- 0.018309} & \num{0.445127 +- 0.018633} & \num{0.492565 +- 0.020417} & \num{0.556986 +- 0.008426} & \num{0.716759 +- 0.006287} \\
 & \gls{gl:ce} \gls{gl:dipa} & \colorblue \bfseries \num{0.215258 +- 0.013370} & \bfseries \num{0.342726 +- 0.036161} & \colorblue \bfseries \num{0.399522 +- 0.043208} & \bfseries \num{0.467969 +- 0.032181} & \colorblue \bfseries \num{0.503135 +- 0.027893} & \colorblue \bfseries \num{0.536371 +- 0.030482} & \bfseries \num{0.601273 +- 0.008712} & \colorblue \bfseries \num{0.719343 +- 0.003934} \\
 & \gls{gl:fl}  & \num{0.177862 +- 0.012051} & \num{0.258456 +- 0.038750} & \num{0.292189 +- 0.024903} & \num{0.340567 +- 0.039914} & \num{0.415502 +- 0.054951} & \num{0.483108 +- 0.036504} & \num{0.540618 +- 0.034931} & \num{0.712539 +- 0.002814} \\
 & \gls{gl:fl} \gls{gl:dipa} & \bfseries \num{0.212308 +- 0.021106} & \colorblue \bfseries \num{0.366818 +- 0.025568} & \bfseries \num{0.363034 +- 0.066010} & \colorblue \bfseries \num{0.470224 +- 0.045804} & \bfseries \num{0.501240 +- 0.026845} & \bfseries \num{0.533612 +- 0.022451} & \colorblue \bfseries \num{0.602266 +- 0.015436} & \bfseries \num{0.714212 +- 0.007882} \\
\bottomrule
\end{tabular}

%% file: images/results_plot_plusf1.tex
\begin{tikzpicture}

\definecolor{indianred}{RGB}{196,78,82}
\definecolor{lavender}{RGB}{234,234,242}
\definecolor{mediumseagreen}{RGB}{85,168,104}
\definecolor{peru}{RGB}{221,132,82}
\definecolor{steelblue}{RGB}{76,114,176}

\pgfplotstableread[col sep = comma]{images/NEW_Eurocrops-Best-Accuracy-Runs_2025-10-29_12-22-01.csv}\tableacc
\pgfplotstableread[col sep = comma]{images/NEW_Eurocrops-Best-CohenKappa-Runs_2025-10-29_12-22-01.csv}\tableck
\pgfplotstableread[col sep = comma]{images/NEW_Eurocrops-Best-F1macro-Runs_2025-11-13_08-50-19.csv}\tablefone

\pgfplotsset{
LinePlot/.style={
    axis background/.style={fill=lavender},
    axis line style={white},
    width=0.95\linewidth,
    height=7cm,
    ymin=0,
    ymax=0.75,
    ytick={0.0,0.25,0.5,0.75},
    yticklabels={0.00,0.25,0.50,0.75},
    ymajorgrids,
    ymajorticks=true,
    tick label style={font=\LARGE},
    y grid style={white},
    ylabel style={font=\LARGE, yshift=0.5cm},
    xtick={1,5,10,20,100,200,500},
    xticklabels={1,5,10,20,100,200,500},
    xlabel style={font=\LARGE, xshift=0.5cm},
    xmin=1,
    xmax=500,
    xmajorticks=true,
    tick align=outside,
    xtick pos=left,
}
}

\begin{groupplot}[
    group style={group name=plot, group size=2 by 3, vertical sep=6em, horizontal sep=10em},
]

\nextgroupplot[
    LinePlot,
    xmode=log,
    log base x=10,
    xlabel=shots $k$,
    xmajorticks=true,
    ylabel=accuracy,
    legend to name=FullLabelBlock,   
    legend columns=4,              
    legend style={/tikz/every even column/.style={column sep=15pt}, font=\LARGE, draw=none, fill=none,
    legend image post style={scale=1.8}},
]
\addplot [forget plot, name path=upper_random_ce, draw=none] table [x=num_samples, y expr=\thisrow{mean_random_ce}+\thisrow{std_random_ce}] {\tableacc};
\addplot [forget plot, name path=lower_random_ce, draw=none] table [x=num_samples, y expr=\thisrow{mean_random_ce}-\thisrow{std_random_ce}] {\tableacc};
\addplot [forget plot, fill=peru!50!white, opacity=0.4] fill between [of=upper_random_ce and lower_random_ce];
\addplot[mark=*, peru, mark size=2pt, line width=1.5] table [x=num_samples, y=mean_random_ce] {\tableacc};

\addplot [forget plot, name path=upper_random_ce_dirichlet, draw=none] table [x=num_samples, y expr=\thisrow{mean_random_ce_dirichlet}+\thisrow{std_random_ce_dirichlet}] {\tableacc};
\addplot [forget plot, name path=lower_random_ce_dirichlet, draw=none] table [x=num_samples, y expr=\thisrow{mean_random_ce_dirichlet}-\thisrow{std_random_ce_dirichlet}] {\tableacc};
\addplot [forget plot, fill=indianred!50!white, opacity=0.4] fill between [of=upper_random_ce_dirichlet and lower_random_ce_dirichlet];
\addplot[mark=diamond*, indianred, mark size=2pt, line width=1.5] table [x=num_samples, y=mean_random_ce_dirichlet] {\tableacc};

\addplot [forget plot, name path=upper_random_fcl, draw=none] table [x=num_samples, y expr=\thisrow{mean_random_fcl}+\thisrow{std_random_fcl}] {\tableacc};
\addplot [forget plot, name path=lower_random_fcl, draw=none] table [x=num_samples, y expr=\thisrow{mean_random_fcl}-\thisrow{std_random_fcl}] {\tableacc};
\addplot [forget plot, fill=steelblue!50!white, opacity=0.4] fill between [of=upper_random_fcl and lower_random_fcl];
\addplot[mark=square*, steelblue, mark size=2pt, line width=1.5] table [x=num_samples, y=mean_random_fcl] {\tableacc};

\addplot [forget plot, name path=upper_random_fcl_dirichlet, draw=none] table [x=num_samples, y expr=\thisrow{mean_random_fcl_dirichlet}+\thisrow{std_random_fcl_dirichlet}] {\tableacc};
\addplot [forget plot, name path=lower_random_fcl_dirichlet, draw=none] table [x=num_samples, y expr=\thisrow{mean_random_fcl_dirichlet}-\thisrow{std_random_fcl_dirichlet}] {\tableacc};
\addplot [forget plot, fill=mediumseagreen!50!white, opacity=0.4] fill between [of=upper_random_fcl_dirichlet and lower_random_fcl_dirichlet];
\addplot[mark=triangle*, mediumseagreen, mark size=2pt, line width=1.5] table [x=num_samples, y=mean_random_fcl_dirichlet] {\tableacc};

\legend{
  \gls{gl:ce},
  \gls{gl:ce} \gls{gl:dipa},
  \gls{gl:fl},
  \gls{gl:fl} \gls{gl:dipa}
}

\nextgroupplot[
    LinePlot,
    xmode=log,
    log base x=10,
    xlabel=shots $k$,
    xmajorticks=true,
]
\addplot [forget plot, name path=upper_pretrain_ce, draw=none] table [x=num_samples, y expr=\thisrow{mean_pretrain_ce}+\thisrow{std_pretrain_ce}] {\tableacc};
\addplot [forget plot, name path=lower_pretrain_ce, draw=none] table [x=num_samples, y expr=\thisrow{mean_pretrain_ce}-\thisrow{std_pretrain_ce}] {\tableacc};
\addplot [forget plot, fill=peru!50!white, opacity=0.4] fill between [of=upper_pretrain_ce and lower_pretrain_ce];
\addplot[mark=*, peru, mark size=2pt, line width=1.5] table [x=num_samples, y=mean_pretrain_ce] {\tableacc};

\addplot [forget plot, name path=upper_pretrain_ce_dirichlet, draw=none] table [x=num_samples, y expr=\thisrow{mean_pretrain_ce_dirichlet}+\thisrow{std_pretrain_ce_dirichlet}] {\tableacc};
\addplot [forget plot, name path=lower_pretrain_ce_dirichlet, draw=none] table [x=num_samples, y expr=\thisrow{mean_pretrain_ce_dirichlet}-\thisrow{std_pretrain_ce_dirichlet}] {\tableacc};
\addplot [forget plot, fill=indianred!50!white, opacity=0.4] fill between [of=upper_pretrain_ce_dirichlet and lower_pretrain_ce_dirichlet];
\addplot[mark=diamond*, indianred, mark size=2pt, line width=1.5] table [x=num_samples, y=mean_pretrain_ce_dirichlet] {\tableacc};

\addplot [forget plot, name path=upper_pretrain_fcl, draw=none] table [x=num_samples, y expr=\thisrow{mean_pretrain_fcl}+\thisrow{std_pretrain_fcl}] {\tableacc};
\addplot [forget plot, name path=lower_pretrain_fcl, draw=none] table [x=num_samples, y expr=\thisrow{mean_pretrain_fcl}-\thisrow{std_pretrain_fcl}] {\tableacc};
\addplot [forget plot, fill=steelblue!50!white, opacity=0.4] fill between [of=upper_pretrain_fcl and lower_pretrain_fcl];
\addplot[mark=square*, steelblue, mark size=2pt, line width=1.5] table [x=num_samples, y=mean_pretrain_fcl] {\tableacc};

\addplot [forget plot, name path=upper_pretrain_fcl_dirichlet, draw=none] table [x=num_samples, y expr=\thisrow{mean_pretrain_fcl_dirichlet}+\thisrow{std_pretrain_fcl_dirichlet}] {\tableacc};
\addplot [forget plot, name path=lower_pretrain_fcl_dirichlet, draw=none] table [x=num_samples, y expr=\thisrow{mean_pretrain_fcl_dirichlet}-\thisrow{std_pretrain_fcl_dirichlet}] {\tableacc};
\addplot [forget plot, fill=mediumseagreen!50!white, opacity=0.4] fill between [of=upper_pretrain_fcl_dirichlet and lower_pretrain_fcl_dirichlet];
\addplot[mark=triangle*, mediumseagreen, mark size=2pt, line width=1.5] table [x=num_samples, y=mean_pretrain_fcl_dirichlet] {\tableacc};

\nextgroupplot[
    LinePlot,
    xmode=log,
    log base x=10,
    xlabel=shots $k$,
    xmajorticks=true,
    ylabel=Cohen's kappa,
]

\addplot [forget plot, name path=upper_random_ce, draw=none] table [x=num_samples, y expr=\thisrow{mean_random_ce}+\thisrow{std_random_ce}] {\tableck};
\addplot [forget plot, name path=lower_random_ce, draw=none] table [x=num_samples, y expr=\thisrow{mean_random_ce}-\thisrow{std_random_ce}] {\tableck};
\addplot [forget plot, fill=peru!50!white, opacity=0.4] fill between [of=upper_random_ce and lower_random_ce];
\addplot[mark=*, peru, mark size=2pt, line width=1.5] table [x=num_samples, y=mean_random_ce] {\tableck};

\addplot [forget plot, name path=upper_random_ce_dirichlet, draw=none] table [x=num_samples, y expr=\thisrow{mean_random_ce_dirichlet}+\thisrow{std_random_ce_dirichlet}] {\tableck};
\addplot [forget plot, name path=lower_random_ce_dirichlet, draw=none] table [x=num_samples, y expr=\thisrow{mean_random_ce_dirichlet}-\thisrow{std_random_ce_dirichlet}] {\tableck};
\addplot [forget plot, fill=indianred!50!white, opacity=0.4] fill between [of=upper_random_ce_dirichlet and lower_random_ce_dirichlet];
\addplot[mark=diamond*, indianred, mark size=2pt, line width=1.5] table [x=num_samples, y=mean_random_ce_dirichlet] {\tableck};

\addplot [forget plot, name path=upper_random_fcl, draw=none] table [x=num_samples, y expr=\thisrow{mean_random_fcl}+\thisrow{std_random_fcl}] {\tableck};
\addplot [forget plot, name path=lower_random_fcl, draw=none] table [x=num_samples, y expr=\thisrow{mean_random_fcl}-\thisrow{std_random_fcl}] {\tableck};
\addplot [forget plot, fill=steelblue!50!white, opacity=0.4] fill between [of=upper_random_fcl and lower_random_fcl];
\addplot[mark=square*, steelblue, mark size=2pt, line width=1.5] table [x=num_samples, y=mean_random_fcl] {\tableck};

\addplot [forget plot, name path=upper_random_fcl_dirichlet, draw=none] table [x=num_samples, y expr=\thisrow{mean_random_fcl_dirichlet}+\thisrow{std_random_fcl_dirichlet}] {\tableck};
\addplot [forget plot, name path=lower_random_fcl_dirichlet, draw=none] table [x=num_samples, y expr=\thisrow{mean_random_fcl_dirichlet}-\thisrow{std_random_fcl_dirichlet}] {\tableck};
\addplot [forget plot, fill=mediumseagreen!50!white, opacity=0.4] fill between [of=upper_random_fcl_dirichlet and lower_random_fcl_dirichlet];
\addplot[mark=triangle*, mediumseagreen, mark size=2pt, line width=1.5] table [x=num_samples, y=mean_random_fcl_dirichlet] {\tableck};

\nextgroupplot[
    LinePlot,
    xmode=log,
    log base x=10,
    xlabel=shots $k$,
    xmajorticks=true,
]
\addplot [forget plot, name path=upper_pretrain_ce, draw=none] table [x=num_samples, y expr=\thisrow{mean_pretrain_ce}+\thisrow{std_pretrain_ce}] {\tableck};
\addplot [forget plot, name path=lower_pretrain_ce, draw=none] table [x=num_samples, y expr=\thisrow{mean_pretrain_ce}-\thisrow{std_pretrain_ce}] {\tableck};
\addplot [forget plot, fill=peru!50!white, opacity=0.4] fill between [of=upper_pretrain_ce and lower_pretrain_ce];
\addplot[mark=*, peru, mark size=2pt, line width=1.5] table [x=num_samples, y=mean_pretrain_ce] {\tableck};

\addplot [forget plot, name path=upper_pretrain_ce_dirichlet, draw=none] table [x=num_samples, y expr=\thisrow{mean_pretrain_ce_dirichlet}+\thisrow{std_pretrain_ce_dirichlet}] {\tableck};
\addplot [forget plot, name path=lower_pretrain_ce_dirichlet, draw=none] table [x=num_samples, y expr=\thisrow{mean_pretrain_ce_dirichlet}-\thisrow{std_pretrain_ce_dirichlet}] {\tableck};
\addplot [forget plot, fill=indianred!50!white, opacity=0.4] fill between [of=upper_pretrain_ce_dirichlet and lower_pretrain_ce_dirichlet];
\addplot[mark=diamond*, indianred, mark size=2pt, line width=1.5] table [x=num_samples, y=mean_pretrain_ce_dirichlet] {\tableck};

\addplot [forget plot, name path=upper_pretrain_fcl, draw=none] table [x=num_samples, y expr=\thisrow{mean_pretrain_fcl}+\thisrow{std_pretrain_fcl}] {\tableck};
\addplot [forget plot, name path=lower_pretrain_fcl, draw=none] table [x=num_samples, y expr=\thisrow{mean_pretrain_fcl}-\thisrow{std_pretrain_fcl}] {\tableck};
\addplot [forget plot, fill=steelblue!50!white, opacity=0.4] fill between [of=upper_pretrain_fcl and lower_pretrain_fcl];
\addplot[mark=square*, steelblue, mark size=2pt, line width=1.5] table [x=num_samples, y=mean_pretrain_fcl] {\tableck};

\addplot [forget plot, name path=upper_pretrain_fcl_dirichlet, draw=none] table [x=num_samples, y expr=\thisrow{mean_pretrain_fcl_dirichlet}+\thisrow{std_pretrain_fcl_dirichlet}] {\tableck};
\addplot [forget plot, name path=lower_pretrain_fcl_dirichlet, draw=none] table [x=num_samples, y expr=\thisrow{mean_pretrain_fcl_dirichlet}-\thisrow{std_pretrain_fcl_dirichlet}] {\tableck};
\addplot [forget plot, fill=mediumseagreen!50!white, opacity=0.4] fill between [of=upper_pretrain_fcl_dirichlet and lower_pretrain_fcl_dirichlet];
\addplot[mark=triangle*, mediumseagreen, mark size=2pt, line width=1.5] table [x=num_samples, y=mean_pretrain_fcl_dirichlet] {\tableck};

\nextgroupplot[
    LinePlot,
    xmode=log,
    log base x=10,
    xlabel=shots $k$,
    ymax=0.250,
    ytick={0.0, 0.05, 0.10, 0.15, 0.20, 0.25},
    yticklabels={0.0, 0.05, 0.10, 0.15, 0.20, 0.25},
    xmajorticks=true,
    ylabel=macro-averaged F1,
]

\addplot [forget plot, name path=upper_random_ce, draw=none] table [x=num_samples, y expr=\thisrow{mean_random_ce}+\thisrow{std_random_ce}] {\tablefone};
\addplot [forget plot, name path=lower_random_ce, draw=none] table [x=num_samples, y expr=\thisrow{mean_random_ce}-\thisrow{std_random_ce}] {\tablefone};
\addplot [forget plot, fill=peru!50!white, opacity=0.4] fill between [of=upper_random_ce and lower_random_ce];
\addplot[mark=*, peru, mark size=2pt, line width=1.5] table [x=num_samples, y=mean_random_ce] {\tablefone};

\addplot [forget plot, name path=upper_random_ce_dirichlet, draw=none] table [x=num_samples, y expr=\thisrow{mean_random_ce_dirichlet}+\thisrow{std_random_ce_dirichlet}] {\tablefone};
\addplot [forget plot, name path=lower_random_ce_dirichlet, draw=none] table [x=num_samples, y expr=\thisrow{mean_random_ce_dirichlet}-\thisrow{std_random_ce_dirichlet}] {\tablefone};
\addplot [forget plot, fill=indianred!50!white, opacity=0.4] fill between [of=upper_random_ce_dirichlet and lower_random_ce_dirichlet];
\addplot[mark=diamond*, indianred, mark size=2pt, line width=1.5] table [x=num_samples, y=mean_random_ce_dirichlet] {\tablefone};

\addplot [forget plot, name path=upper_random_fcl, draw=none] table [x=num_samples, y expr=\thisrow{mean_random_fcl}+\thisrow{std_random_fcl}] {\tablefone};
\addplot [forget plot, name path=lower_random_fcl, draw=none] table [x=num_samples, y expr=\thisrow{mean_random_fcl}-\thisrow{std_random_fcl}] {\tablefone};
\addplot [forget plot, fill=steelblue!50!white, opacity=0.4] fill between [of=upper_random_fcl and lower_random_fcl];
\addplot[mark=square*, steelblue, mark size=2pt, line width=1.5] table [x=num_samples, y=mean_random_fcl] {\tablefone};

\addplot [forget plot, name path=upper_random_fcl_dirichlet, draw=none] table [x=num_samples, y expr=\thisrow{mean_random_fcl_dirichlet}+\thisrow{std_random_fcl_dirichlet}] {\tablefone};
\addplot [forget plot, name path=lower_random_fcl_dirichlet, draw=none] table [x=num_samples, y expr=\thisrow{mean_random_fcl_dirichlet}-\thisrow{std_random_fcl_dirichlet}] {\tablefone};
\addplot [forget plot, fill=mediumseagreen!50!white, opacity=0.4] fill between [of=upper_random_fcl_dirichlet and lower_random_fcl_dirichlet];
\addplot[mark=triangle*, mediumseagreen, mark size=2pt, line width=1.5] table [x=num_samples, y=mean_random_fcl_dirichlet] {\tablefone};

\nextgroupplot[
    LinePlot,
    xmode=log,
    log base x=10,
    xlabel=shots $k$,
    ymax=0.250,
    ytick={0.0, 0.05, 0.10, 0.15, 0.20, 0.25},
    yticklabels={0.0, 0.05, 0.10, 0.15, 0.20, 0.25},
    xmajorticks=true,
]

\addplot [forget plot, name path=upper_pretrain_ce, draw=none] table [x=num_samples, y expr=\thisrow{mean_pretrain_ce}+\thisrow{std_pretrain_ce}] {\tablefone};
\addplot [forget plot, name path=lower_pretrain_ce, draw=none] table [x=num_samples, y expr=\thisrow{mean_pretrain_ce}-\thisrow{std_pretrain_ce}] {\tablefone};
\addplot [forget plot, fill=peru!50!white, opacity=0.4] fill between [of=upper_pretrain_ce and lower_pretrain_ce];
\addplot[mark=*, peru, mark size=2pt, line width=1.5] table [x=num_samples, y=mean_pretrain_ce] {\tablefone};

\addplot [forget plot, name path=upper_pretrain_ce_dirichlet, draw=none] table [x=num_samples, y expr=\thisrow{mean_pretrain_ce_dirichlet}+\thisrow{std_pretrain_ce_dirichlet}] {\tablefone};
\addplot [forget plot, name path=lower_pretrain_ce_dirichlet, draw=none] table [x=num_samples, y expr=\thisrow{mean_pretrain_ce_dirichlet}-\thisrow{std_pretrain_ce_dirichlet}] {\tablefone};
\addplot [forget plot, fill=indianred!50!white, opacity=0.4] fill between [of=upper_pretrain_ce_dirichlet and lower_pretrain_ce_dirichlet];
\addplot[mark=diamond*, indianred, mark size=2pt, line width=1.5] table [x=num_samples, y=mean_pretrain_ce_dirichlet] {\tablefone};

\addplot [forget plot, name path=upper_pretrain_fcl, draw=none] table [x=num_samples, y expr=\thisrow{mean_pretrain_fcl}+\thisrow{std_pretrain_fcl}] {\tablefone};
\addplot [forget plot, name path=lower_pretrain_fcl, draw=none] table [x=num_samples, y expr=\thisrow{mean_pretrain_fcl}-\thisrow{std_pretrain_fcl}] {\tablefone};
\addplot [forget plot, fill=steelblue!50!white, opacity=0.4] fill between [of=upper_pretrain_fcl and lower_pretrain_fcl];
\addplot[mark=square*, steelblue, mark size=2pt, line width=1.5] table [x=num_samples, y=mean_pretrain_fcl] {\tablefone};

\addplot [forget plot, name path=upper_pretrain_fcl_dirichlet, draw=none] table [x=num_samples, y expr=\thisrow{mean_pretrain_fcl_dirichlet}+\thisrow{std_pretrain_fcl_dirichlet}] {\tablefone};
\addplot [forget plot, name path=lower_pretrain_fcl_dirichlet, draw=none] table [x=num_samples, y expr=\thisrow{mean_pretrain_fcl_dirichlet}-\thisrow{std_pretrain_fcl_dirichlet}] {\tablefone};
\addplot [forget plot, fill=mediumseagreen!50!white, opacity=0.4] fill between [of=upper_pretrain_fcl_dirichlet and lower_pretrain_fcl_dirichlet];
\addplot[mark=triangle*, mediumseagreen, mark size=2pt, line width=1.5] table [x=num_samples, y=mean_pretrain_fcl_dirichlet] {\tablefone};

\end{groupplot}

\node[
  anchor=north,
  below=3cm of plot c2r3.south,
  xshift=-9cm, 
  align=center
]{\pgfplotslegendfromname{FullLabelBlock}};

\node[below = 2cm of plot c1r3.south] {\LARGE (a) randomly initialized network};
\node[below = 2cm of plot c2r3.south] {\LARGE (b) pretrained network};

\end{tikzpicture}

%% file: main.bbl
\begin{thebibliography}{xx}

\bibitem[Alem and Kumar, 2022]{Alem22:transferlearningLCLU}
Alem, A., Kumar, S., 2022.
 Transfer learning models for land cover and land use classification in remote
  sensing image.
 {\em Applied Artificial Intelligence}, 36(1), 2014192.

\bibitem[Chen et al., 2019]{Chen18:fewshotclassification}
Chen, W.-Y., Liu, Y.-C., Kira, Z., Wang, Y.-C.~F., Huang, J.-B., 2019.
 A closer look at few-shot classification.
 \emph{International Conference on Learning Representations}.

\bibitem[Claverie et al., 2025]{Claverie25:ECM}
Claverie, M., Chan, A.~X., Ramos, H., Koeble, R., K{\"o}rner, M., See, L., {van
  der Velde}, M., 2025.
 EuroCrops 2.0: Multi-annual harmonized parcel level crop type data
  cross-linked to European union-wide survey, statistical, and Earth
  observation products.

\bibitem[Daum\'{e}~III and Marcu, 2005]{Daume05:DirichletClustering}
Daum\'{e}~III, H., Marcu, D., 2005.
 A Bayesian model for supervised clustering with the Dirichlet process prior.
 6, 1551–1577.

\bibitem[Finn et al., 2017]{Finn17:MAML}
Finn, C., Abbeel, P., Levine, S., 2017.
 Model-agnostic meta-learning for fast adaptation of deep networks.
 \emph{Proceedings of the 34th International Conference on Machine Learning
  (ICML)}, ~70, JMLR.org, 1126--1135.

\bibitem[Kluger et al., 2021]{Kluger21:FPSA}
Kluger, D.~M., Wang, S., Lobell, D.~B., 2021.
 Two shifts for crop mapping: Leveraging aggregate crop statistics to improve
  satellite-based maps in new regions.
 {\em Remote Sensing of Environment}, 262, 112488.

\bibitem[Kurian et al., 2024]{Vinija24:transferlearning}
Kurian, V., Jacob, V., Kuruvilla, J., 2024.
 Approach of transfer learning in remote sensing image classification.
 \emph{2024 1st International Conference on Trends in Engineering Systems and
  Technologies (ICTEST)}, 1--3.

\bibitem[Lin et al., 2020]{Lin18:FocalLoss}
Lin, T.-Y., Goyal, P., Girshick, R., He, K., Dollár, P., 2020.
 Focal loss for dense object detection.
 {\em IEEE Transactions on Pattern Analysis and Machine Intelligence}, 42(2),
  318-327.

\bibitem[Lipton et al., 2018]{Lipton18:BlackBox}
Lipton, Z., Wang, Y.-X., Smola, A., 2018.
 Detecting and correcting for label shift with black box predictors.
 J.~Dy, A.~Krause (eds), \emph{Proceedings of the 35th International Conference
  on Machine Learning (ICML)}, Proceedings of Machine Learning Research, ~80,
  PMLR, 3122--3130.

\bibitem[Mohammadi et al., 2024]{Mohammadi24:fewshotcropmappingdirichlet}
Mohammadi, S., Belgiu, M., Stein, A., 2024.
 Few-shot learning for crop mapping from satellite image time series.
 {\em Remote Sensing}, 16(6).

\bibitem[Ochal et al., 2023]{Ochael23:classimbalanceFSL}
Ochal, M., Patacchiola, M., Vazquez, J., Storkey, A., Wang, S., 2023.
 Few-shot learning with class imbalance.
 {\em IEEE Transactions on Artificial Intelligence}, 4(5), 1348-1358.

\bibitem[Qi et al., 2023]{QI23:deeplcropclassification}
Qi, Y., Bitelli, G., Mandanici, E., Trevisiol, F., 2023.
 {Application of deep learning crop classification model based on multispectral
  and SAR satellite imagery}.
 {\em The International Archives of the Photogrammetry, Remote Sensing and
  Spatial Information Sciences}, XLVIII-1/W2-2023, 1515--1521.

\bibitem[Rademacher and Doroslovački, 2021]{Rademacher21:Dirichletpriors}
Rademacher, P., Doroslovački, M., 2021.
 Bayesian learning for regression using dirichlet prior distributions of
  varying localization.
 \emph{2021 IEEE Statistical Signal Processing Workshop (SSP)}, 236--240.

\bibitem[Raghu et al., 2019]{Raghu19:ANIL}
Raghu, A., Raghu, M., Bengio, S., Vinyals, O., 2019.
 Rapid learning or feature reuse?
 \emph{International Conference on Learning Representations (ICLR)}.

\bibitem[Reuss et al., 2025a]{Reuss25:MTLSSL}
Reuss, J., Macdonald, J., Becker, S., Gikalo, E., Schultka, K., Richter, L.,
  K{\"o}rner, M., 2025a.
 Benchmarking for practice: Few-shot time-series crop-type classification on
  the \textsc{EuroCropsML} dataset.

\bibitem[Reuss et al., 2025b]{Reuss25:EML}
Reuss, J., Macdonald, J., Becker, S., Richter, L., K{\"o}rner, M., 2025b.
 {EuroCropsML}: A time series benchmark dataset for few-shot crop type
  classification.
 {\em Nature Scientific Data}.

\bibitem[Rouba and Larabi, 2023]{Rouba23:heterogenoustransferlearning}
Rouba, M., Larabi, M. E.~A., 2023.
 Improving remote sensing classification with transfer learning: Exploring the
  impact of heterogenous transfer learning.
 {\em Engineering Proceedings}, 56(1).

\bibitem[Ru{ß}wurm et al., 2020]{Russwurm20:MTL}
Ru{ß}wurm, M., Wang, S., Körner, M., Lobell, D., 2020.
 Meta-learning for few-shot land cover classification.
 \emph{IEEE/CVF Conference on Computer Vision and Pattern Recognition Workshops
  (CVPRW)}, 9.

\bibitem[Saini and Ghosh, 2018]{Saini18:cropclassificationRFSVM}
Saini, R., Ghosh, S.~K., 2018.
 Crop classification on single date Sentinel-2 imagery using random forest and
  support vector machine.
 {\em The International Archives of the Photogrammetry, Remote Sensing and
  Spatial Information Sciences}, XLII-5, 683--688.

\bibitem[Schneider et al., 2021]{Schneider2021:TinyEuroCrops}
Schneider, M., Broszeit, A., K{\"o}rner, M., 2021.
 {EuroCrops}: A pan-european dataset for time series crop type classification.
 P.~Soille, S.~Loekken, S.~Albani (eds), \emph{Conference on Big Data from
  Space (BiDS)}, Publications Office of the European Union.

\bibitem[Schneider and Körner, 2021]{Schneider21:SIT}
Schneider, M., Körner, M., 2021.
 [Re] Satellite image time series classification with pixel-set encoders and
  temporal self-attention.
 7.

\bibitem[Schneider et al., 2023]{Schneider2023:EuroCrops}
Schneider, M., Schelte, T., Schmitz, F., K{\"o}rner, M., 2023.
 EuroCrops: The largest harmonized open crop dataset across the European Union.
 {\em Scientific Data}, 10(1), 612.

\bibitem[Sipka et al., 2022]{Sipka22:hitchiker}
Sipka, T., Sulc, M., Matas, J., 2022.
 The hitchhiker’s guide to prior-shift adaptation.
 \emph{2022 IEEE/CVF Winter Conference on Applications of Computer Vision
  (WACV)}, IEEE Computer Society, 2031--2039.

\bibitem[Steck and Jaakkola, 2002]{Harald02:DirichletBayes}
Steck, H., Jaakkola, T., 2002.
 On the dirichlet prior and bayesian regularization.
 S.~Becker, S.~Thrun, K.~Obermayer (eds), \emph{Advances in Neural Information
  Processing Systems}, ~15, MIT Press.

\bibitem[Tseng et al., 2022]{Tseng22:TIML}
Tseng, G., Kerner, H.~R., Rolnick, D., 2022.
 Timl: Task-informed meta-learning for agriculture.

\bibitem[Tseng et al., 2021]{Tseng20:CropHarvest}
Tseng, G., Zvonkov, I., Nakalembe, C., Kerner, H.~R., 2021.
 {CropHarvest}: A global dataset for crop-type classification.
 J.~Vanschoren, S.~Yeung (eds), \emph{Proceedings of the Neural Information
  Processing Systems (NeurIPS) Track on Datasets and Benchmarks}, ~1, 14.

\bibitem[Vaswani et al., 2017]{vaswani_vanillatransformer}
Vaswani, A., Shazeer, N., Parmar, N., Uszkoreit, J., Jones, L., Gomez, A.~N.,
  Kaiser, {\L}., Polosukhin, I., 2017.
 Attention is all you need.
 \emph{Advances in Neural Information Processing Systems (NeurIPS)}, ~30, 15.

\bibitem[Veilleux et al., 2021]{Veilleux21:dirichletevaluation}
Veilleux, O., Boudiaf, M., Piantanida, P., Ben~Ayed, I., 2021.
 Realistic evaluation of transductive few-shot learning.
 \emph{Advances in Neural Information Processing Systems}, ~34, Curran
  Associates, Inc., 9290--9302.

\bibitem[Wang et al., 2020]{Wang20:MTL}
Wang, S., Ru{ß}wurm, M., Körner, M., Lobell, D.~B., 2020.
 Meta-learning for few-shot time series classification.
 \emph{IEEE International Geoscience and Remote Sensing Symposium (IGARSS)},
  7041--7044.

\end{thebibliography}
